\documentclass[10pt,twocolumn,letterpaper]{article}

\usepackage{cvm}
\usepackage{times}
\usepackage{epsfig}
\usepackage{graphicx}
\usepackage{amsmath}
\usepackage{amssymb}

\usepackage[pagebackref=true,breaklinks=true,letterpaper=true,colorlinks,bookmarks=false]{hyperref}

\cvmfinalcopy

\def\cvmPaperID{****} 

\ifcvmfinal\fi
\begin{document}

\title{Effect of Instance Normalization on Fine-Grained Control for Sketch-Based Face Image Generation}

\author{Zhihua Cheng\\
NEL-BITA, University of Science and Technology of China\\
{\tt\small czh666@mail.ustc.edu.cn}
\and
Xuejin Chen*\\
NEL-BITA, University of Science and Technology of China\\
{\tt\small xjchen99@ustc.edu.cn}
}

\maketitle

\begin{abstract}
    Sketching is an intuitive and effective way for content creation. While significant progress has been made for photorealistic image generation by using generative adversarial networks, it remains challenging to take a fine-grained control on synthetic content.  
    The instance normalization layer, which is widely adopted in existing image translation networks, washes away details in the input sketch and leads to loss of precise control on the desired shape of the generated face images.
    In this paper, we comprehensively investigate the effect of instance normalization on generating photorealistic face images from hand-drawn sketches.
    We first introduce a visualization approach to analyze the feature embedding for sketches with a group of specific changes.  
    Based on the visual analysis, we modify the instance normalization layers in the baseline image translation model. 
    We elaborate a new set of hand-drawn sketches with 11 categories of specially designed changes and conduct extensive experimental analysis.  
    The results and user studies demonstrate that our method markedly improve the quality of synthesized images and the conformance with user intention.
\end{abstract}

\section{Introduction}

Photorealistic face image synthesis from hand-drawn sketches has drawn a lot of attention in computer graphics and computer vision for many years. Typical approaches use generative and adversarial networks~(GANs)~\cite{gan}, that  stack convolutional, normalization and nonlinearity layers as generators. Normalization layers normalize the parameter distribution in order to alleviate the issue of slow convergence in gradient update process and avoid vanishing gradient and exploding gradient problem, which is vastly important in GANs.

Many normalization layers have been developed in recent GANs for various goals, such as batch normalization~\cite{bn}, group normalization~\cite{gn}, layer normalization~\cite{ln} and instance normalization~\cite{instance_norm}. 
Batch normalization~\cite{bn} eliminates the influence of internal covariate shift, effectively avoids the possible problems of gradient vanishing and gradient exploding in the process of gradient backpropagation, and speeds up the training time. 
Group normalization~\cite{gn} organizes the channels of a layer into different groups, and computes the mean and standard deviation within each group independently for normalization. It is independent of batch size, thus it is frequently used in tasks which prefers small mini-batch size, such as object detection and video classification.
Layer normalization~\cite{ln} computes the mean and variance used for normalization over all the channels of a single layer. It is more suitable for recurrent neural networks. Unlike batch normalization, layer normalization~\cite{instance_norm} performs exactly the same computation at training and test times.

Instance normalization is similar to layer normalization but goes one step further. It computes the mean and variance for normalization over \emph{each} channel in \emph{each} training example.
Recent studies show that instance normalization performs well on visual tasks such as style transfer and image translation~\cite{pix2pixhd,spade,cyclegan} when replacing batch normalization in GANs architecture. 
Nonetheless, instance normalization layers tend to wash away detailed information conveyed by the input sketches, thus it results in descent of the feature expression ability and imprecise control on face generation. 

It is essential to support fine-grained control in sketch-based content creation. 
We comprehensively investigate the effect of instance normalization on fine-grained control in sketch-based photorealistic face image generation using data visualization methods. 
We utilize principal component analysis (PCA)~\cite{pca} to visualize and analyze features extracted by the generator from sketches. 
Consequently, we propose to remove the first two instance normalization layers in the baseline generator, and show that this simple modification in the generator results in a significant improvement on control accuracy in image generation. 
We conduct extensive experiments and interactive user studies to evaluate our proposed method, and the results demonstrate that our method surpasses the baseline method on image quality and control precision on the sketch-to-image task.

\section{Related Work}
\subsection{Image-to-Image Translation}
Image-to-image translation aims to convert an input image from one domain to another given the input-output image pairs as training data, in other words, to generate corresponding image according to the input image while the two images share the same scene structure. At present, many researchers employ adversarial manner to train deep neural networks in image-to-image translation tasks~\cite{cyclegan,bicyclegan,spagan,munit,crn,cfgan,sis,cfgan,maskgan}.

The concept of image-to-image translation was first proposed by pix2pix~\cite{pix2pix}, derived from generative adversarial networks while conditioned on images. 
The pix2pix network consists of a generator $G$ and a discriminator $D$. The generator converts the input image from a source domain to a target domain, and the discriminator tells the generated images apart from real images. 
This model can be applied to a variety of image translation scenarios, such as lable maps to streetscapes, edge maps to photos, image colorization, and so on.
However, the original pix2pix model has limits of low resolution images, at most $256 \times 256$. 
When pix2pix is applied to generate images with a higher resolution, the training process will be unstable and the generation quality will decline dramatically.
In order to improve the resolution of synthesized images, a subsequent model, pix2pixHD~\cite{pix2pixhd}, is proposed to generate images from semantic lable maps by a coarse-to-fine generator and a multiscale discriminator. 
However, the instance normalization layers used in pix2pixHD tend to wash away semantic information.
In order to efficiently preserve and propagate semantic information throughout the network, GauGAN~\cite{spade} utilizes semantic segmentation masks to modulate activations in the normalization layers through a spatially-adaptive and learned transformation. 
These models can also be applied to edge-to-photo generation when conditioned on edge map and photo pairs. 
However, the large gap between synthesized edge maps and hand-drawn sketches challenges the generalization ability of these models.
It inspires us to investigate the effect of normalization layers more deeply on information propagation in the network architecture for sketch-based face image synthesis.

\subsection{Normalization Layers}
In current deep neural networks, normalization layers play an important role for stabilizing the training process. 
We introduce several common normalization methods in detail. 
Batch normalization~\cite{bn} is a method that normalizes activations in a network across the mini-batch. 
It calculates the mean and variance among one channel over each mini-batch. Then, it learns two parameters to scale and shift the normalized activations.
Batch Normalization provides a strong way to reduce internal covariant shift problem and speed up the training process.
Group normalization~\cite{gn} divides the channels of activations into groups and then calculates the mean and standard deviation over the channel groups of each training sample for normalization.
Group normalization is frequently adopted in tasks such as object detection, semantic segmentation and video classification. It helps deep learning models work better at small mini-batch size.
Layer normalization~\cite{ln} computes the mean and variance used for normalization over all the channels of a single layer. It is more suitable for recurrent neural networks. 
Unlike batch normalization, layer normalization performs exactly the same computation at training and test times.
Instance normalization~\cite{instance_norm} is similar to batch normalization. The only difference is that batch normalization computes the mean and variance among a mimi-batch, while instance normalization operates across only one channel of a single layer. 
Instance normalization performs well on style transfer~\cite{carigan,apdrawinggan,stylization,cartoongan,singlegan,transgaga,harmonic} tasks when replacing batch normalization in GANs.

\section{Feature Embedding Visualization}
We investigate the effect of instance normalization on fine-grained control in sketch-based photorealistic image generation by designing a set of hand-drawn sketches and visualizing the feature embedding.
In Sec.~\ref{sec:pix2pixhd}, we review our baseline method pix2pixHD.
In Sec.~\ref{sec:visualize}, we visualize the feature embedding from hand-drawn sketches by the generator of pix2pixHD, and analyze the visualization results to investigate the effect of instance normalization. 
Based on the analysis, we introduce our design on the generator network for sketch-based face generation in Sec.~\ref{sec:network}.

\subsection{The pix2pixHD Baseline}\label{sec:pix2pixhd}
Pix2pixHD~\cite{pix2pixhd} is an image translation model based on conditional GAN.
It adopts an improved adversarial loss and the network architecture to generate high-quality and high-resolution images from input semantic label maps. 
A coarse-to-fine generator and a multiscale discriminator are introduced to increase the image resolution and enhance texture details.
Using this model, more realistic images in dimension of $2048 \times 1024$ can be generated. 

Its generator $G$ is composed of two sub-networks, a global generator network $G_1$ and a local enhance network $G_2$. 
$G_1$ is used to generate a base image, then $G_2$ is used to increase the image resolution with texture details. 
In order to distinguish real and generated images with high resolution, the discriminator requires a larger receptive field. 
Therefore, pix2pixHD uses a multi-scale discriminator, which is composed of three discriminators $D_1$, $D_2$ and $D_3$ in three scales, to preserve both global and local information. 
The loss function of this model is composed of three parts: adversarial loss $L_{GAN}(G,D_k)$, feature matching loss $L_{FM}(G,D_k)$, and VGG perceptual loss $L_{VGG}(G)$. The full objective is formulated as:
\begin{equation}
\begin{split}
&\underset{G}{\min}\Bigg(\bigg(\underset{D_1,D_2,D_3}{\max} \sum_{k=1,2,3}{L_{GAN}\big(G,D_k\big)}\bigg)+  \\
&\lambda \bigg(\frac{1}{3}\sum_{k=1,2,3}{L_{FM}\big(G,D_k\big)}+L_{VGG}{\big(G\big)}\bigg)\Bigg),
\end{split}
\end{equation}
\noindent
where $\lambda$ controls the importance of the three terms.

\begin{figure}
	\centering
	\includegraphics[width=\columnwidth]{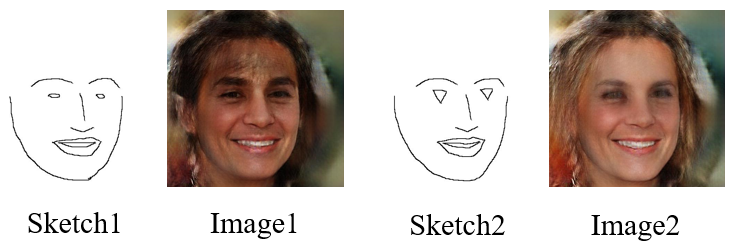}
	\caption{Pix2pixHD fails to support fine-grained control on the synthesized images when the input sketch changes locally at the eye shape.}
	\label{fig:problem-in}
\end{figure}

Pix2pixHD can be applied to sketch-based face generation when trained using pairs of synthesized contours and photos. 
We developed an interactive system to support users to create face images by sketching. 
However, when an user tries to change the shape of a local part, pix2pixHD tends to change the entire image globally, as shown in Fig.~\ref{fig:problem-in}.
Therefore, we investigate the network architecture and feature embedding through a comprehensive experiment.

\begin{figure*}[htbp]
	\centering
	\includegraphics[width=\textwidth]{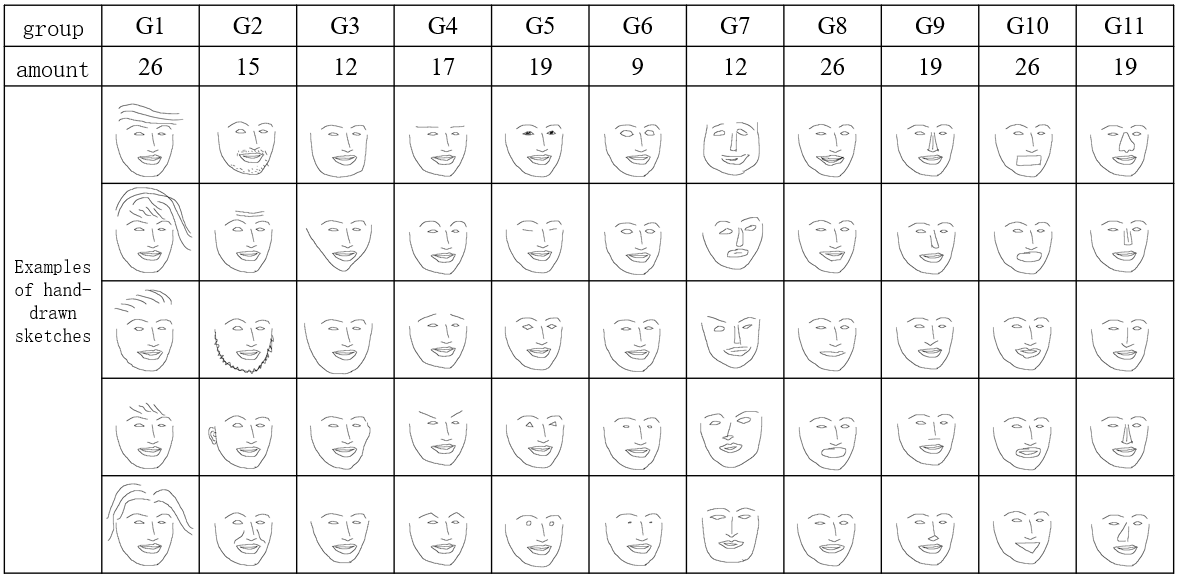}
	\caption{We collect 11 groups of hand-drawn sketches with specific changes at each group. G1: Add hair; G2: Add new attributes, such as whiskers, wrinkles, and ears; G3: Face shape change; G4: Eyebrow change; G5: Eye shape change; G6: Eye size change; G7: Graffiti-drawn; G8: Mouth shape change; G9: Nose shape change; G10: Mouth shape change with the same eyes as G9; G11: Nose shape change with the same eyes as G8. Specifically, there is no correlation between G7 and the other 10 groups. Except G7, in the sketches of other groups, only a particular area or attribute is modified while the rest sketches in this group remain the same. While sketches in G8 and G11 have the same eyes, sketches in G9 and G10 have the same eyes, the eye shapes in other groups are different.}
	\label{fig:hand_drawn_contours}
\end{figure*}

\begin{figure}[htbp]
	\begin{center}
		\includegraphics[width=0.2 \textwidth]{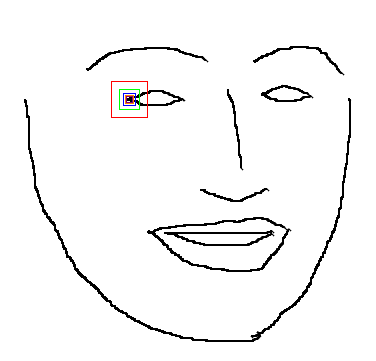}
	\end{center}
	\caption{The receptive fields of corresponding left eye corner points in the feature maps of $L_0\sim L_4$. The smallest purple box corresponds to $L_0$, while the biggest red box corresponds to $L_4$. }
	\label{fig:receptive}
\end{figure}

\subsection{Feature Visualization}\label{sec:visualize}


In order to analyze how the image translation model extracts the face shape features consistent with the user's intention for the hand-drawn sketches that typically have little details and geometric deformation, we collect 11 groups of sketches with changes in a specific area in each group. It contains 198 sketches totally of resolution $512\times512$. Fig.~\ref{fig:hand_drawn_contours} shows some examples of the 11 groups of sketches. And we refer to the 11 groups as $G1,\ldots,G11$.

We refer to the first five layers of the global generator of pix2pixHD as $L_0,\ldots,L_4$. 
With a sketch fed into the generator, five groups of feature maps can be obtained from $L_0$ to $L_4$.
For each point in a layer, we can extract a $k$-dimensional feature vector, where $k$ is the channel number of the feature maps in one layer.
We select the corner of the left eye for visual analysis. 
The five feature vectors are denoted as $\boldsymbol{v}_0,\boldsymbol{v}_1,\boldsymbol{v}_2,\boldsymbol{v}_3,\boldsymbol{v}_4$ in dimensions of 48, 96, 192, 384 and 768, respectively. 
The receptive fields of the five vectors are $7\times7$, $9\times9$, $13\times13$, $21\times21$, and $37\times37$ respectively on the input sketch, as shown in Fig.~\ref{fig:receptive}. 

\begin{figure*}[htb]
	\centering
	\includegraphics[width=0.9 \textwidth]{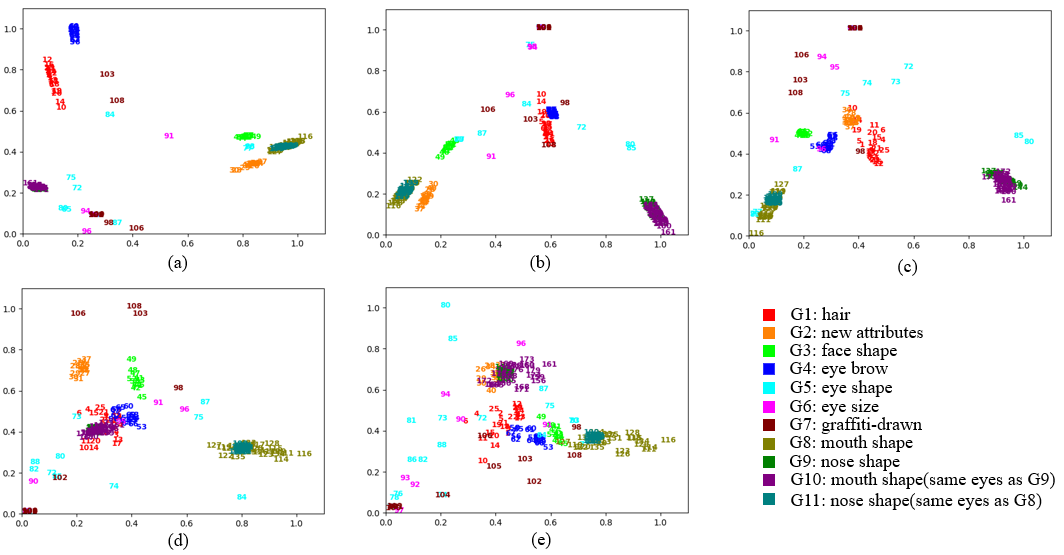}
	\caption{Visualization of feature embedding $\boldsymbol{v}_0\sim\boldsymbol{v}_4$ in the first five layers of the global generator of pix2pixHD for the left eye corner for 11 groups of sketches.  }
	\label{fig:pca_0}
\end{figure*}

\paragraph{Visualization with PCA.} Principal component analysis~(PCA)~\cite{pca} is a widely-used dimension reduction method and retains the statistic characteristics of data in high dimensional space. 
We employ PCA on $\boldsymbol{v}_0,\ldots, \boldsymbol{v}_4$ to map the high-dimensional vectors into 2D vectors and plot them in a 2D space. 
Fig.~\ref{fig:pca_0} show the visualization results with PCA on $\boldsymbol{v}_0,\ldots,\boldsymbol{v}_4$. 
In group 1, all the sketches share the same eye shape but with different hair styles. They are supposed to have the same feature embedding for the left eye corner at the early convolutional layers. However, as shown in Fig.~\ref{fig:pca_0}, the feature vectors scatter across a wide range (red numbers). 
Similarly, though the eye strokes do not change in the sketches of the same group, the feature embedding of G2 (orange), G3 (lime), G4 (blue), G8 (olive), G9 (green), G10 (purple) and G11 (teal) scatter.
%
This phenomenon indicates that the instance normalization in each layer takes the sketch information outside of its corresponding receptive field into account and then affects the feature embedding of a local region. 
By comparing visualization results of $\boldsymbol{v}_0\sim\boldsymbol{v}_4$ comprehensively, we find out that the feature vectors belonging to the groups without eye changes, such as G1, G2, G3, G4, G8, G9, G10, and G11, distribute more and more dispersedly within each group from layer 1 to layer 4, indicating that with more instance normalization, changing other parts on the input sketch has more and more influence on the local feature embedding, resulting in an awful change of eyes in the generated images.
Consequently, we modify the normalization in the generator of pix2pixHD to keep local shape details. 

\begin{figure*}[htb]
	\centering
	\includegraphics[width=0.9 \textwidth]{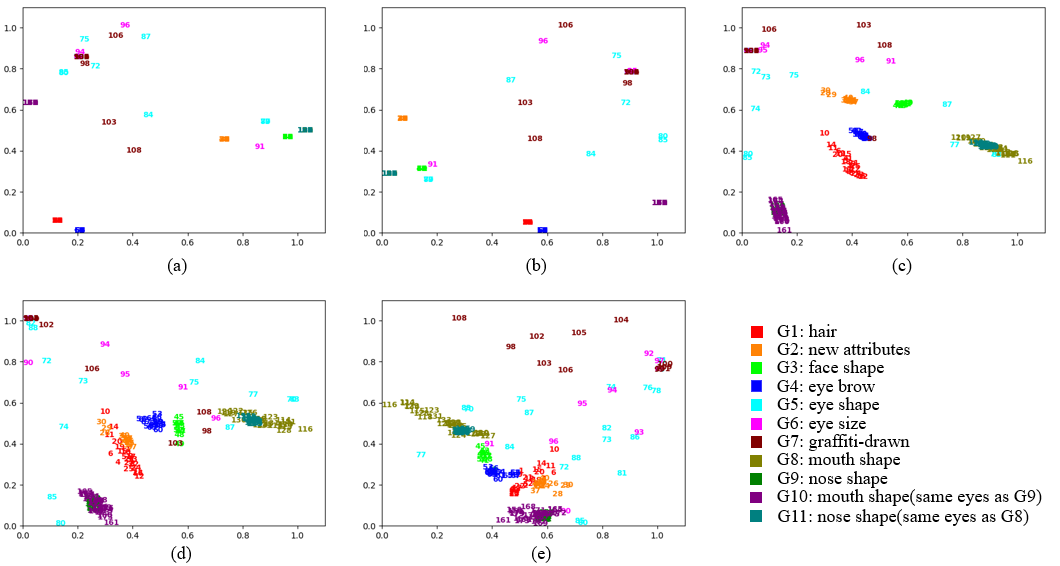}
	\caption{Visualization of feature embedding $\boldsymbol{v}_0^{'}\sim\boldsymbol{v}_4^{'}$ in the first five layers of our generator for the left eye corner for 11 groups of sketches.}
	\label{fig:pca_1}
\end{figure*}

\section{Our Network Design}\label{sec:network}
With instance normalization in the baseline generator, the activation of convolutional output is normalized in a channel-wise manner and then modulated with unified scale and bias within each channel. 
This operation, to a certain extent, leads to a negative effect that a local change in the input sketch broadcasts globally, resulting in a degraded capacity of fine-grained control on the generated images.
However, the vanishing gradient or exploding gradient problem is bound to emerge when the instance normalization layers in the feature embedding stage of the generator is abandoned, making the training process difficult to converge. 

\begin{figure}[htbp]
	\centering
	\includegraphics[width=\columnwidth]{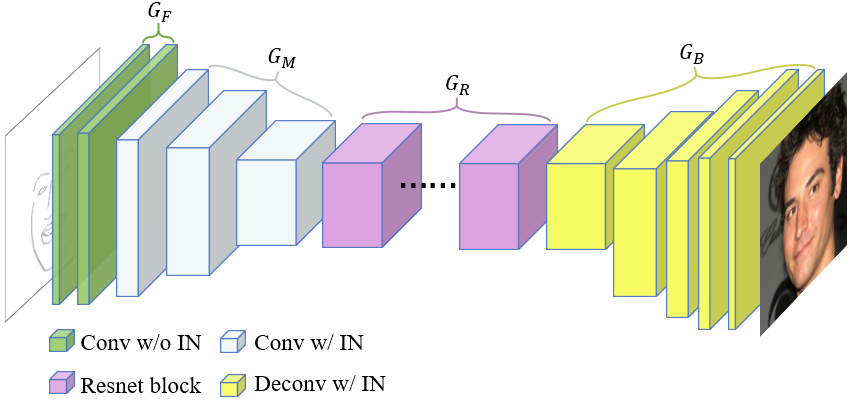}
	\caption{Network architecture of our generator. }
	\label{fig:our_generator}
\end{figure}

Based on the consideration above, we only remove the first two instance normalization layers in the global generator of pix2pixHD and keep the rest the same as pix2pixHD. 
The architecture of our generator is shown in Fig.~\ref{fig:our_generator}. 
It consists of four components: a convolutional front-end without normalization $G_F$, a down-sampled convolutional mid-end $G_M$, a set of residual blocks $G_R$, and a transposed convolutional back-end $G_B$.  
$G_F$, which is composed of two unnormalized convolution and activation layers, embeds local shape information of the input sketch.
In other words, the local shape information of the sketch can flow through the network without spatially broadcasting, thus effectively supports fine-grained control on generated images. 

We use the same three-scale discriminator with pix2pixHD. 
Each discriminator is built on the PatchGAN~\cite{pix2pix} architecture. 
At each scale, the input sketch is concatenated with the corresponding face images, resized and fed into the corresponding discriminator.

To evaluate the superiority of our proposed method at the level of feature embedding, we conducted the visual analysis in the same way as described in Sec.~\ref{sec:visualize}.
We refer to the five feature vectors in our generator as $\boldsymbol{v}_0^{'},\boldsymbol{v}_1^{'},\boldsymbol{v}_2^{'},\boldsymbol{v}_3^{'},\boldsymbol{v}_4^{'}$.
Fig.~\ref{fig:pca_1} (a) and Fig.~\ref{fig:pca_1} (b) demonstrate the results of PCA visualization on $\boldsymbol{v}_0^{'}$ and $\boldsymbol{v}_1^{'}$. 
The feature vectors from G1, G2, G3, and G4 are located at the same point in each group. The feature vectors from G8 and G11 are gathered on the same point since these two groups share the same eye shape, so do the feature vectors from G9 and G10.
On the contrary, the feature vectors from G5, G6, and G7 distribute dispersedly, indicating that after the removal of instance normalization in the first two layers of generator, the feature embedding of the left eye corner conforms with the local shape of the input sketches inside of the corresponding receptive fields. 
The sketch change within the receptive field will influence the embedded features of the corresponding point but not outside of the receptive field.

We visualize the feature embedding of $\boldsymbol{v}_2^{'},\boldsymbol{v}^{'}_3, \boldsymbol{v}_4^{'}$ in 
Fig.~\ref{fig:pca_1} (c) (d) (e). 

Compare with the PCA visualization results of $\boldsymbol{v}_2,\boldsymbol{v}_3,\boldsymbol{v}_4$ in 
Fig.~\ref{fig:pca_0} (c) (d) (e), the $\boldsymbol{v}_2, \boldsymbol{v}_3, \boldsymbol{v}_4$ belonging to groups without eye shape change, such as G1, G2, G3, G4, G8, G9, G10, and G11, have a bigger in-class distance compared with $\boldsymbol{v}_2^{'}, \boldsymbol{v}_3^{'}, \boldsymbol{v}_4^{'}$. 
This indicates that the removal of the first two instance normalization layers in pix2pixHD generator can better extract the low-level features of the input sketch and keep local shape information. 
The improvement on generator network can effectively alleviate the issue that changing one part of the sketch influences the other parts of the generated image. 
Therefore, our network better supports fine-grained control and enhances texture details on the generated images.

\section{Experiments on Sketch-based Face Generation}

To evaluate the effectiveness of our method on interactive face synthesis by sketching, we conducted extensive experiments with a wide range of handdrawn sketches and local editing. 
We develop an interactive interface for drawing sketches and displaying the generated photos in real time.

\paragraph{Datasets.}
CelebA-HQ dataset contains 30k face images in resolution $1024\times1024$. All the face images are cropped and globally aligned according to face landmarks.
To produce paired training data of sketches and face images, we extracted 68 landmarks from each face image in the CelebA-HQ dataset, connect these landmark points in sequence, and draw lines in width of two pixels to synthesize contours as sketches.
Those contours are more simple and clean than other types of synthetic sketches like edge maps~\cite{csagan} and mask edge maps~\cite{maskgan}. 
In our experiments, all the synthetic contour images and face images are resized into $512\times512$. 
After removing the photos for which landmark extraction fails, the training set contains 14,973 pairs of contour image and face photo, while the test set contains 4,992 pairs of contour and face photo. 

\subsection{Data Augmentation}\label{sec:augmentation} 
\begin{figure}[htb]
	\centering
	\includegraphics[width=0.18 \textwidth]{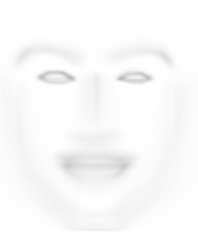}
	\caption{The average face of all the synthesized contours from the face photos. }
	\label{fig:average_face}
\end{figure}

The face photos of the CalebA-HQ dataset are precisely aligned based on facial landmarks. Fig.~\ref{fig:average_face} shows an average face of all the synthesized contours. 
It shows that the facial features and facial contours of training data are basically in the same position, in other words, the training data is global aligned. 
This issue leads to degraded generalization ability of the model.

\begin{figure}[htb]
	\centering
	\includegraphics[width=0.35 \textwidth]{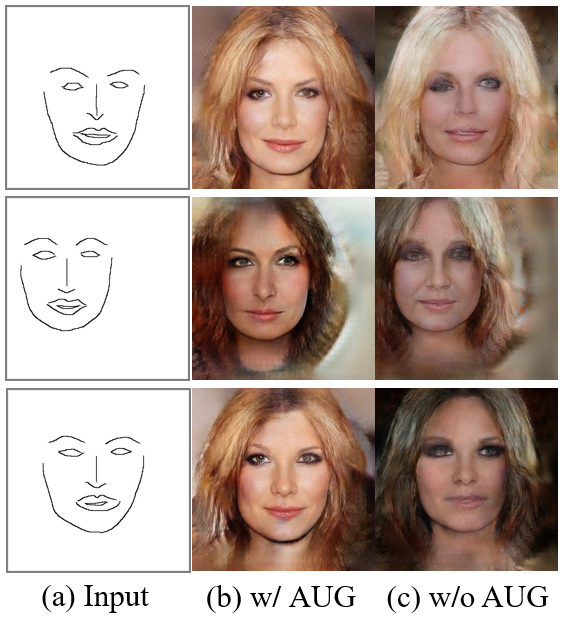}
	\caption{Side-by-side comparison between images generated w/ and w/o data augmentation. }
	\label{fig:data_augmentation}
\end{figure}

In order to imitate the human hand-drawn sketches, we apply random translation and rotation to the training contour images. 
Specifically, offsets randomly selected from $[-d,d]^2$ and angles randomly selected from $[-\theta,\theta]$ are applied to the training contours, where $d$ is the maximum offset and $\theta$ is the maximum angle.
We set $d=25$ and $\theta=7^\circ$ in our experiments. 
However, the training face photos are not translated or rotated because we expect the generated images to remain global aligned regardless of the spatial location of the input sketches.
Fig.~\ref{fig:data_augmentation} illustrates the comparative results between images generated before data augmentation and that after data augmentation. 
The quality of generated images after data augmentation is greatly improved compared with those results without data augmentation when the input sketches deviate from the training examples spatially.

\subsection{Qualitative Comparisons}\label{sec:results}
We conduct extensive experiments on different model with groups of elaborated hand-drawn sketches to verify the effect of instance normalization in generator on sketch-based face photo generation.

\begin{figure}[htb]
	\centering
	\includegraphics[width=\columnwidth ]{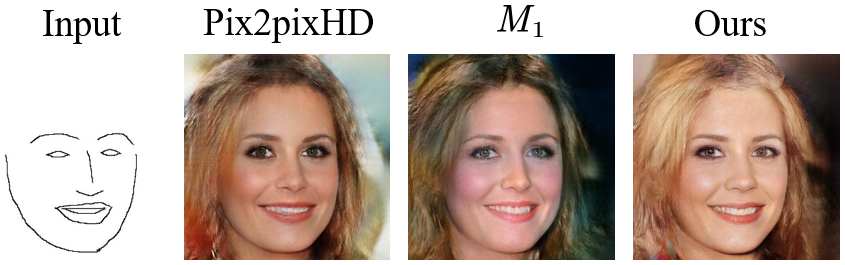}
	\caption{Comparison between results generated by different models with different amount of instance normalization layers. }
	\label{fig:ablation}
\end{figure}

To further explore how the amount of instance normalization layers in generator network affect the sketch-based face image generation quality, we train another model which takes the generator of pix2pixHD got rid of the first five instance normalization layers as its own generator. 
This model is refered to as $M_1$ for convenience.
We compare the face images generated by our model, $M_1$ and pix2pixHD. 
As you can see in Fig.~\ref{fig:ablation}, our model generates the most photorealistic images in contrast to the $M_1$ model which generates images lacking realism. 
It indicates that removing too many instance normalization layers in generator can weaken the model because training process is difficult to converge without normalization.
So it's reasonable for our model just to remove the first two instance normalization layers in generator. 

\begin{figure}[htb]
	\centering
	\includegraphics[width=0.45 \textwidth]{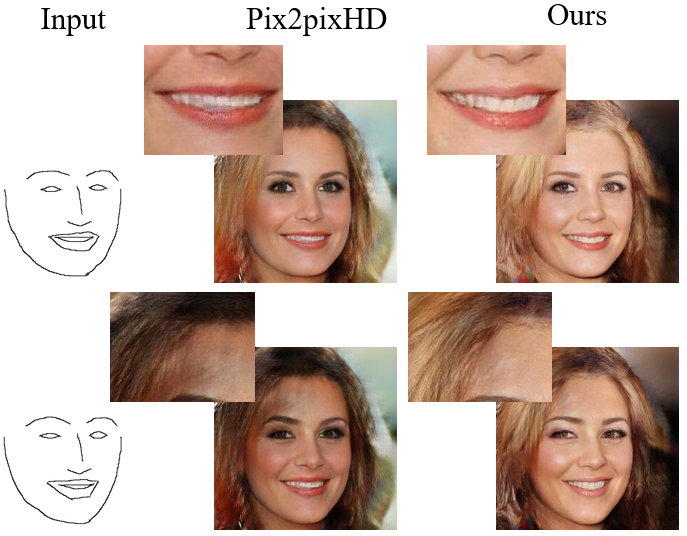}
	\caption{Face images generated by our model and the baseline model with different mouth shape. The top row shows that the textures of teeth generated by pix2pixHD baseline are blurry, while our results have more realistic textures at the teeth area. The bottom row shows that some chaotic noises often emerge in the forehead area of the image generated by pix2pixHD baseline because the strokes of the mouth affect the other regions during feature embedding. In comparison, our model produces more globally realistic results.}
	\label{fig:compare_1}
\end{figure}

Then we perform comparative experiments between our model and pix2pixHD baseline on hand-drawn sketches.
Results shown in Fig.~\ref{fig:compare_1} demonstrate that the baseline model frequently fails to synthesize realistic textures and many images generated by baseline model exhibit blurry artifacts.
We ascribe this issue to instance normalization that the generator creates blurry artifact to dominate the statistics to fool instance normalization layer. 
In contrast, the blurry artifacts alleviate obviously in our results when the first two instance normalization steps are removed from the generator.
Meanwhile, our results are more plausible with fine-grained textures due to our generator without the first two instance norm layers preserving more underlying information of input sketches.

\begin{figure}[htb]
	\centering
	\includegraphics[width=0.4 \textwidth]{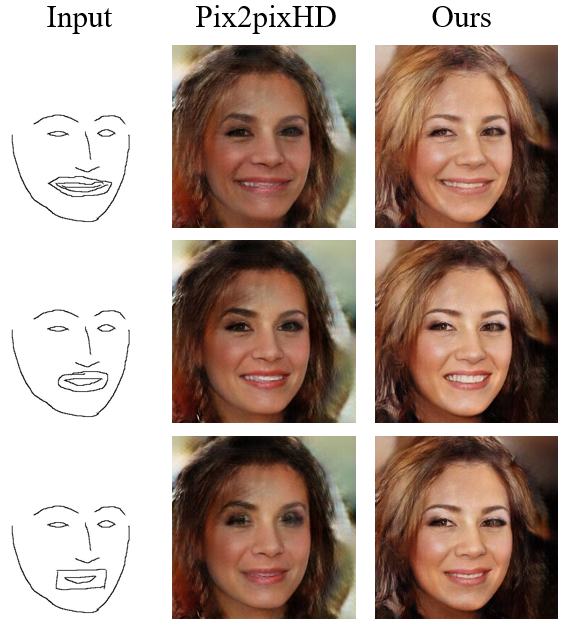}
	\caption{Comparison between our model and baseline model tested with mouth-altered sketches. }
	\label{fig:compare_2}
\end{figure}

\begin{figure}[htb]
	\centering
	\includegraphics[width=0.45 \textwidth]{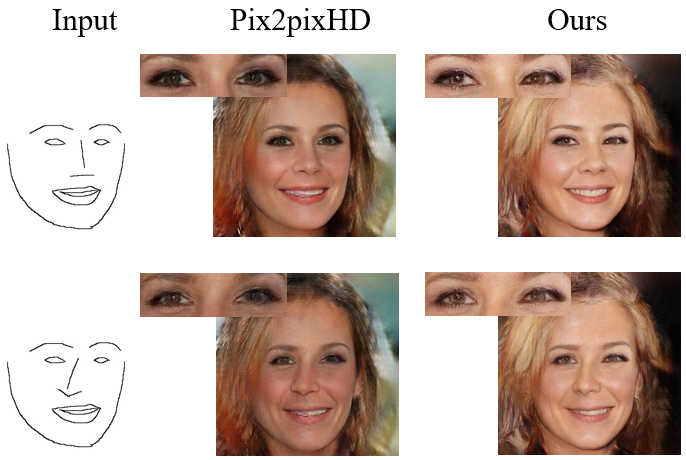}
	\caption{Comparison between our model and baseline model when the input sketches change locally at the nose shape.}
	\label{fig:compare_3}
\end{figure}

Our model can achieve fine-grained control on generated images. 
When we modify the strokes to represent different face attributes or change the overall face shapes in the input freehand sketches, the corresponding parts of images generated by our model change consistently while other areas remain unchanged. 
In comparison, using the baseline pix2pixHD model, modifying local strokes of sketches influences not only the content in corresponding areas but also the content in other areas in the generated images. 
Fig.~\ref{fig:compare_2} and Fig.~\ref{fig:compare_3}~ show several face images generated by our model and baseline model when changing lines of mouth and nose separately in sketches.
The results shown in first and second row of Fig.~\ref{fig:compare_2} demonstrate that the images generated by our model change obviously in mouth shape and preserve structure conformance with input sketches when modifying the lines of mouth in sketches. 
But the results generated by baseline model do not have an obvious change in mouth shape.
The results shown in Fig.~\ref{fig:compare_3} illustrate that the images generated by our model remain unchanged in other areas especially in eyes when altering the shapes of nose in sketches, but the images generated by baseline model change obviously in eyes direction. 
And as shown in the third row of Fig.~\ref{fig:compare_2}, the generation results of our model do not change except the mouth when modifying the mouth shape in sketches, while the generation quality of baseline model is degraded vastly especially in the eye area. 

\begin{table}[h]
	\centering	
	\caption{Results of user study.}
	\begin{tabular}{|c|c|c|}\hline
		& Baseline~\cite{pix2pixhd} & Ours \\\hline
		Preferance & $12.8\%$ & $\textbf{87.2\%}$\\\hline
	\end{tabular}
	\label{tab:user_study}
\end{table} 

Finally, user study is conducted to evaluate the perceptual generation quality of our model in comparison with baseline model. 
We organize 60 volunteers to join our experiment, each of them is tested with about 100 trials. The users are supposed to select images that are more realistic and match the input sketches better.  
The results reported in Table~\ref{tab:user_study} indicates that compared with baseline model, face images generated by our model are more plausible and controllable according to the test users. 

\section{Conclusion}
In this paper, we investigate the feature embedding in state-of-the-art image translation model on the sketch-based face image generation task.
We collect 11 groups of sketches with specific designs and utilize PCA to visually analyze the feature embedding for sketches.
The analysis indicates that the instance normalization tends to wash away the local shape information in the input sketches.
We improve the baseline image translation model by modifying the instance normalization layers in its generator.
The modified model effectively conveys fine-grained shape information through the image translation model and produces photorealistic face images that conforms with the input sketches on both local shape and global structure.
Extensive experiments demonstrate the effectiveness of our proposed method on the image quality and conformance with user intention in a sketch-based face image generation system.


{\small
\bibliographystyle{cvm}
\bibliography{cvmbib}

\begin{thebibliography}{10}\itemsep=-1pt

\bibitem{ln}
J.~L. Ba, J.~R. Kiros, and G.~E. Hinton.
\newblock Layer normalization, 2016.

\bibitem{carigan}
K.~Cao, J.~Liao, and L.~Yuan.
\newblock Carigans: Unpaired photo-to-caricature translation, 2018.

\bibitem{cartoongan}
Y.~Chen, Y.-K. Lai, and Y.-J. Liu.
\newblock Cartoongan: Generative adversarial networks for photo cartoonization.
\newblock In {\em Proceedings of the IEEE Conference on Computer Vision and
  Pattern Recognition (CVPR)}, June 2018.

\bibitem{crn}
{Chen Qifeng} and {Koltun Vladlen}.
\newblock Photographic image synthesis with cascaded refinement networks.
\newblock In {\em \textit{The IEEE International Conference on Computer Vision
  (ICCV)}}, Oct 2017.

\bibitem{pca}
{Citation. Hotelling H. }.
\newblock Analysis of a complex of statistical variables into principal
  components.
\newblock {\em \textit{Journal of Educational Psychology}}, 24(6):417--441,
  1933.

\bibitem{spagan}
H.~Emami, M.~M. Aliabadi, M.~Dong, and R.~B. Chinnam.
\newblock Spa-gan: Spatial attention gan for image-to-image translation, 2020.

\bibitem{gan}
{Goodfellow Ian}, {Pouget-Abadie Jean}, {Mirza Mehdi}, {Xu Bing},
  D.~Warde-Farley, S.~Ozair, A.~Courville, and Y.~Bengio.
\newblock Generative adversarial nets.
\newblock In {\em \textit{Advances in Neural Information Processing
  Systems(NIPS)}}, pages 2672--2680. 2014.

\bibitem{sis}
{Hao Dong}, {Simiao Yu}, {Chao Wu}, and {Yike Guo}.
\newblock Semantic image synthesis via adversarial learning.
\newblock In {\em \textit{{IEEE} International Conference on Computer
  Vision(ICCV)}}, pages 5707--5715, 2017.

\bibitem{munit}
X.~Huang, M.-Y. Liu, S.~Belongie, and J.~Kautz.
\newblock Multimodal unsupervised image-to-image translation.
\newblock In {\em ECCV}, 2018.

\bibitem{pix2pix}
{Isola Phillip}, {Zhu Jun-Yan}, {Zhou Tinghui}, and {Efros Alexei A.}
\newblock Image-to-image translation with conditional adversarial networks.
\newblock In {\em \textit{The IEEE Conference on Computer Vision and Pattern
  Recognition (CVPR)}}, July 2017.

\bibitem{cyclegan}
{Jun{-}Yan Zhu}, {Taesung Park}, {Phillip Isola}, and {Alexei A. Efros}.
\newblock Unpaired image-to-image translation using cycle-consistent
  adversarial networks.
\newblock In {\em \textit{International Conference on Computer Vision(ICCV)}},
  pages 2242--2251, 2017.

\bibitem{maskgan}
{Lee, Cheng-Han}, {Liu, Ziwei}, {Wu, Lingyun}, and {Luo, Ping}.
\newblock Maskgan: Towards diverse and interactive facial image manipulation.
\newblock In {\em \textit{IEEE Conference on Computer Vision and Pattern
  Recognition (CVPR)}}, 2020.

\bibitem{csagan}
{Li, Yuhang}, {Chen, Xuejin}, {Wu, Feng}, and {Zha, Zheng-Jun}.
\newblock Linestofacephoto: Face photo generation from lines with conditional
  self-attention generative adversarial networks.
\newblock In {\em \textit{Proceedings of the 27th ACM International Conference
  on Multimedia}}, pages 2323--2331, 2019.

\bibitem{stylization}
A.~Sanakoyeu, D.~Kotovenko, S.~Lang, and B.~Ommer.
\newblock A style-aware content loss for real-time hd style transfer.
\newblock In {\em Proceedings of the European Conference on Computer Vision
  (ECCV)}, September 2018.

\bibitem{bn}
{Sergey Ioffe} and {Christian Szegedy}.
\newblock Batch normalization: Accelerating deep network training by reducing
  internal covariate shift.
\newblock In {\em \textit{International Conference on Machine Learning(ICML)}},
  volume~37, pages 448--456, 2015.

\bibitem{spade}
{Taesung Park}, {Ming{-}Yu Liu}, {Ting{-}Chun Wang}, and {Jun{-}Yan Zhu}.
\newblock Semantic image synthesis with spatially-adaptive normalization.
\newblock In {\em \textit{Conference on Computer Vision and Pattern
  Recognition(CVPR)}}, pages 2337--2346, 2019.

\bibitem{cfgan}
{Takuhiro Kaneko}, {Kaoru Hiramatsu}, and {Kunio Kashino}.
\newblock Generative attribute controller with conditional filtered generative
  adversarial networks.
\newblock In {\em \textit{Conference on Computer Vision and Pattern
  Recognition(CVPR)}}, pages 7006--7015, 2017.

\bibitem{instance_norm}
{Ulyanov Dmitry}, {Vedaldi Andrea}, and {Lempitsky Victor}.
\newblock Improved texture networks: Maximizing quality and diversity in
  feed-forward stylization and texture synthesis.
\newblock In {\em \textit{The IEEE Conference on Computer Vision and Pattern
  Recognition (CVPR)}}, July 2017.

\bibitem{pix2pixhd}
{Wang Ting-Chun}, {Liu Ming-Yu}, {Zhu Jun-Yan}, {Tao Andrew}, {Kautz Jan}, and
  {Catanzaro Bryan}.
\newblock High-resolution image synthesis and semantic manipulation with
  conditional gans.
\newblock In {\em \textit{The IEEE Conference on Computer Vision and Pattern
  Recognition (CVPR)}}, June 2018.

\bibitem{transgaga}
W.~Wu, K.~Cao, C.~Li, C.~Qian, and C.~C. Loy.
\newblock Transgaga: Geometry-aware unsupervised image-to-image translation.
\newblock In {\em CVPR}, 2019.

\bibitem{apdrawinggan}
R.~Yi, Y.-J. Liu, Y.-K. Lai, and P.~L. Rosin.
\newblock Apdrawinggan: Generating artistic portrait drawings from face photos
  with hierarchical gans.
\newblock In {\em Proceedings of the IEEE/CVF Conference on Computer Vision and
  Pattern Recognition (CVPR)}, June 2019.

\bibitem{singlegan}
X.~Yu, X.~Cai, Z.~Ying, T.~Li, and G.~Li.
\newblock Singlegan: Image-to-image translation by a single-generator network
  using multiple generative adversarial learning.
\newblock In {\em Asian Conference on Computer Vision}, 2018.

\bibitem{gn}
{Yuxin Wu} and {Kaiming He}.
\newblock Group normalization.
\newblock In {\em \textit{European Conference of Computer Vision(ECCV)}},
  volume 11217, pages 3--19, 2018.

\bibitem{harmonic}
R.~Zhang, T.~Pfister, and J.~Li.
\newblock Harmonic unpaired image-to-image translation.
\newblock In {\em International Conference on Learning Representations}, 2019.

\bibitem{bicyclegan}
J.-Y. Zhu, R.~Zhang, D.~Pathak, T.~Darrell, A.~A. Efros, O.~Wang, and
  E.~Shechtman.
\newblock Toward multimodal image-to-image translation.
\newblock In {\em Advances in Neural Information Processing Systems}, 2017.

\end{thebibliography}
}

\end{document}